\documentclass{article}
\usepackage{spconf,amsmath,graphicx,subfigure,booktabs,float}
\usepackage{array}
\usepackage{bm}
\usepackage{color}
\usepackage{multirow}

\title{A DEEPLY-RECURSIVE CONVOLUTIONAL NETWORK FOR CROWD COUNTING}
\name{Xinghao Ding, Zhirui Lin, Fujin He, Yu Wang, Yue Huang\thanks{This work was supported in part by the National Natural Science Foundation of China under Grants 61571382, 81671766, 61571005, 81671674, U1605252, 61671309  in part by the Guangdong Natural Science Foundation under Grant 2015A030313007, in part by the Fundamental Research Funds for the Central Universities under Grant 20720160075, 20720150169.}}
\address{Fujian Key Laboratory of Sensing and Computing for Smart City, Xiamen University, China\\
School of Information Science and Engineering, Xiamen University, Xiamen, 361005, China\\
*yhuang2010@xmu.edu.cn}
\begin{document}
%
\maketitle
\begin{abstract}
The estimation of crowd count in images has a wide range of applications such as video surveillance, traffic monitoring, public safety and urban planning. Recently, the convolutional neural network (CNN) based approaches have been shown to be more effective in crowd counting than traditional methods that use handcrafted features. However, the existing CNN-based methods still suffer from large number of parameters and large storage space, which require high storage and computing resources and thus limit the real-world application. Consequently, we propose a deeply-recursive network (DR-ResNet) based on ResNet blocks for crowd counting. The recursive structure makes the network deeper while keeping the number of parameters unchanged, which enhances network capability to capture statistical regularities in the context of the crowd. Besides, we generate a new dataset from the video-monitoring data of Beijing bus station. Experimental results have demonstrated that proposed method outperforms most state-of-the-art methods with far less number of parameters.
\end{abstract}
\section{INTRODUCTION}
\label{sec:intro}

The crowd counting tasks aim to estimate the number of humans in the surveillance videos and photos. A single-image crowd counting is very useful in traffic management, disaster prevention and public management. In addition, the technology developed for crowd counting can be applied to  other research areas such as cell microscopes, vehicle counts and so on. Because of its wide range of application, crowd counting is an important research area in the field of computer vision and intelligent video surveillance. As the other computer vision methods, crowd counting is also faced with difficulties in occlusion, scale variances, background clutters, and perspective changes [1].

Nowadays, there are a variety of methods to estimate the crowd count, such as detection-based methods [2-5], regression-based methods [6-8], density estimation based methods [9,10], etc. Most of the initial research is focused on handcrafted features to solve detection or regression tasks. However, these methods have difficulty in handling the scenario with extremely dense crowds and high background clutter. In recent years, the CNN-based methods have been highly praised in this field because their performance is superior to the traditional methods based on handcrafted features [11-13]. The multi-column based network architecture [14], the scale aware model named Hydra CNN [15], and the Crowdnet model [16] all emphasize adaptability to scale changes and can be classified into ensemble methods. Sam et al. [17] trained a Switch-CNN to classify the crowd into three classes depending on crowd density and to select one of 3 regressor networks for actual counting. Similarly, a Contextual Pyramid CNN (CP-CNN) [18] generates the crowd density maps combining global and local contexts obtained by classification of input images and its patches into various density levels. These methods continually improve accuracy of crowd counting, but do not fully consider the problem of limited computing resources in practical applications.

In practical applications, such as safety monitoring, the embedded devices are commonly used. However, the above-mentioned network architectures require complex structures with a big convolution core and many feature maps to capture effective features, resulting in need to store a large amount of parameters. Therefore, these methods require high storage and computing resources, which restricts their application especially in embedded devices whose memory is limited. Therefore, to mitigate the mentioned disadvantages, we proposed a deeply-recursive ResNet architecture named DR-ResNet, which has a small number of model parameters and needs neither multi-staged training nor pre-training.

\begin{figure*}
\centering
\includegraphics[width=17cm]{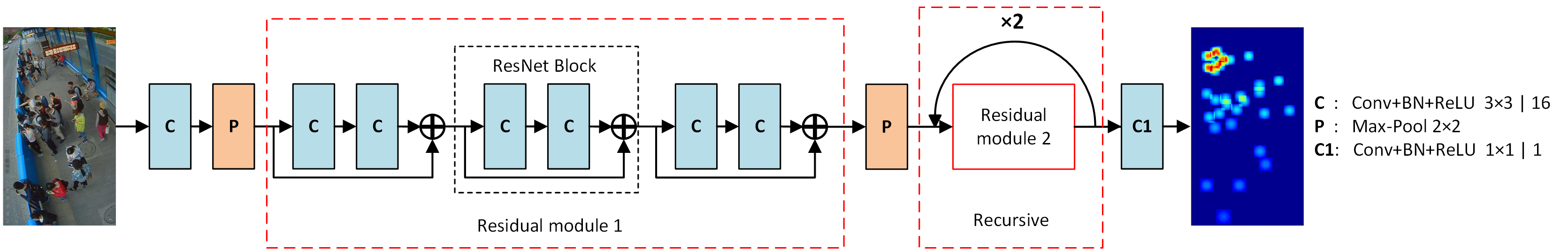}
\caption{The structure of the proposed deeply-recursive network for crowd density map estimation.}
\end{figure*}

The main contributions of this study can be summarized as follows. Due to the network recursiveness and simple structure, the number of parameters in our model is much smaller compared to the existing methods, but our model provides a comparable or even a superior performance on the public crowd counting datasets, which is challenging and representative. Moreover, we introduce a new public scene dataset containing 1,280 images with 16,795 labeled pedestrians for crowd analysis. The introduced dataset is a representative dataset with the most realistic and challenging setting of a crowded scene in intelligent transportation field. Furthermore, we have made our dataset  public\footnote{https://github.com/XMU-smartdsp/Beijing-BRT-dataset}.

The paper is organized as follows. In Section 2, a brief review of crowd counting methods based on crowd density map is provided and the proposed method is described in detail. In Section 3, the obtained experimental results are presented. Lastly, a brief conclusion is given in Section 4.

\section{CROWD COUNTING METHODS}
\label{sec:format}

For a given input image, we first obtain a crowd density map containing a spatial distribution information using our method, and then estimate the number of people by integration [14]. For a head annotation $x_i$ of the image, we represent the annotation as a delta function $\delta(x-x_i)$ and describe its distribution with a Gaussian kernel[9] $G_{\sigma}$ so the density map with $N$ heads is as follows:
\begin{equation}
\begin{array}{l}
F(x)=H(x)*G_{\sigma}(x),H(x)=\sum\limits_{i = 1}^N{\delta(x-x_i)}
\end{array}
\end{equation}
The above method is generally applicable to sparse crowd. For intensive big scene, we generate density map via geometry-adaptive kernels[14], thus it is defined by:
\begin{equation}
F(x)=\sum\limits_{i = 1}^N{\delta(x-x_i)*G_{\sigma_i}},\sigma_i=\beta\bar{d}_i
\end{equation}
where $\sigma_i$ depends on the average distance $\bar{d}_i$ between the head and its nearest k neighbors, while $\beta$ denotes an empirical value and here $\beta=0.3$.

\begin{figure}[h]
\centering
\includegraphics[height=2.2in,width=3.0in,angle=0]{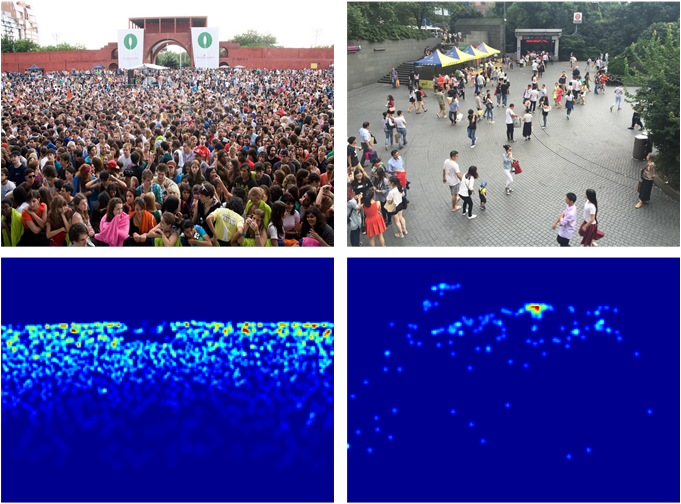}
\caption{The crowd density maps of given crowd images of ShanghaiTech dataset.}
\end{figure}

The model we use is based on ResNet [20] because of its high performance and fast convergence in image classification and regression tasks with a simple network structure. Since the trade-off between the number of network parameters and model performance can be overcome using a recursive strategy, we propose a deep-recursive ResNet (DR-ResNet). The structure of our DR-ResNet is illustrated in Fig. 1. Our ResNet block refers to [21], which consists of 2 convolutional layers and batch normalization using the Rectified Linear Unit (Conv\&BN\&ReLU) structures connected by a shortcut. Each Residual module consists of 3 ResNet blocks. The overall structure contains 2 residual modules. Inspired by [22], we keep the number of parameters belonging to the second residual module constant by weight sharing, while the network becomes deeper. Since the recursive residual module is equivalent to a three-residual-module cascade, the recursive units represent the same mapping function, namely the residual module 2 is represented by $y=f(x)$, so the recursive residual module 2  can be represented by $y=f(f(f(x)))$. Thus, the number of DR-ResNet parameters is equal to that of ResNet-14 (ResNet with 14 convolution layers), but the depth of DR-ResNet is equal to the depth of ResNet-26 (ResNet with 26 convolution layers). Finally we use $1\times1$ convolution as a reconstruction to aggregate the multiple feature maps into a density map. The max pooling method is applied for each $2\times2$ region, so the density map is a quarter of the input image size. The recursive structure makes the network deeper while keeping the number of parameters unchanged, and a deeper structure increases the receptive field of back layers, so there is no need to use a big-size convolution core. Besides, the deeper structure makes the model representation stronger reducing the number of feature maps required of each layer. This greatly reduces the number of model parameters and saves computing resources. The total number of parameters in the proposed model is only 0.028 M(here M denotes $10^6$).

We use Euclidean distance to measure the difference between predicted density map and ground truth of the input patch. The loss function is defined by:
\begin{equation}
L(\Theta)=\frac{1}{2N}\sum\limits_{i = 1}^N{\|F(X_i;\Theta)-F_i\|_2^2}
\end{equation}
where $\Theta$ is a set of  parameters in the DR-ResNet, $N$ is the number of training samples. $L$ denotes the loss between estimated density map $F(X_i;\Theta)$ and ground truth density map $F_i$ of input image $X_i$. The loss is minimized via the batch-based stochastic gradient descent.

\section{EXPERIMENTS}
\label{sec:pagestyle}

In the section, we compare performances of our DR-ResNet model and several state-of-the-art CNN-based methods on three different datasets. Our model is implemented by the Caffe framework developed in [23] using the batch-based stochastic gradient descent (SGD) and backpropagation with the momentum of 0.9, initial learning rate of 0.01 and weight decay of 0.0005.

\subsection{Evaluation on recursive structure}
\label{ssec:subhead}
In the experiment, we used a deep-recursive network with a weight sharing between layers. We also compared the Recursive residual module 1 (R-ResNet) and Recursive residual module 2 (DR-ResNet). Table 1 illustrates that Recursive residual module 1 cannot improve performance but makes it even worse, because the main role of Residual module 1 is to extract the low-level features while the main role of Residual module 2 is to capture the high-level features. Experimental results show that recursive structure should be used at a high level. Besides, the feature maps of Residual module 1 are relatively large, so the Recursive residual module 1 will obviously increase the computational complexity of convolution neural network. Due to the hardware memory limitation, our model uses small feature maps as well as fewer number of parameters. Our goal is to minimize the number of parameters and reduce the computational complexity while maintaining a good performance. Therefore, we used the DR-ResNet illustrated in Fig. 1, which is a good compromise of ResNet-14 and ResNet-26. In Table 2, the number of DR-ResNet parameters is far less than those of other methods, which makes our model more applicable to the embedded devices.

\subsection{Performance comparison}
\label{ssec:subhead}
\subsubsection{Evaluation metrics}
\label{sssec:subhead}
Following the convention of the existing methods [13,14,17], we used the mean absolute error (MAE) and mean squared error (MSE) to evaluate the performance of different methods. The MAE and MSE are defined by:
\begin{equation}
MAE=\frac{1}{N}\sum\limits_{i = 1}^N{|z_i-\hat{z}_i|},MSE=\sqrt{\frac{1}{N}\sum\limits_{i = 1}^N{(z_i-\hat{z}_i)^2}}
\end{equation}
where $N$ is the number of test images, $z_i$ is the actual number of people in the $i^{th}$ image, while $\hat{z}_i$ is the estimated number of people in the $i^{th}$ image. In general, MAE and MSE can respectively indicate the accuracy and robustness. Moreover, we used the number of neural networks parameters (PARAMS) to evaluate computation complexity of a method, which is an important evaluation metric of practical application. When calculating the number of parameters, we assume that the input of network is a three-channel image.

\begin{table}[h]
\caption{Comparison of Recursive residual module 1 (R-ResNet) and Recursive residual module 2 (DR-ResNet).} 
\centering 
\begin{tabular}{c c c c c c} 
\hline
\multirow{2}{*}{Method} &
\multirow{2}{*}{PARAMS} &
\multicolumn{2}{c}{Part\_A} &
\multicolumn{2}{c}{Part\_B} \\
\cline{3-6} & & MAE & MSE & MAE & MSE\\
\hline
ResNet-14 & 0.028M & 93.8 & 137.1 & 18.1 & 27.8\\
ResNet-20 & 0.042M & 90.2 & 136.7 & 15.0 & 22.4\\
ResNet-26 & 0.056M & 85.8 & 128.9 & 14.3 & 20.8\\
R-ResNet & 0.028M & 101.5 & 147.9 & 20.2 & 39.9\\
DR-ResNet & 0.028M & 86.3 & 124.2 & 14.5 & 21.0\\
\hline
\end{tabular}
\label{tab:PPer}
\end{table}

\begin{table*}[ht]
\caption{Comparing the number of model parameters and performances of different methods on Shanghaitech dataset and UCF\_CC\_50 dataest. Bold denotes the best results, and red denotes the second best ones.} 
\centering 
\begin{tabular}{c c c c c c c c} 
\hline 
\multirow{2}{*}{Method} &
\multirow{2}{*}{PARAMS} &
\multicolumn{2}{c}{Part\_A} &
\multicolumn{2}{c}{Part\_B} &
\multicolumn{2}{c}{UCF\_CC\_50} \\
\cline{3-8} & & MAE & MSE & MAE & MSE & MAE & MSE\\
\hline
  Crowdnet[16] &14.7M &- &- &- &- &452.5 &- \\
  Hydra 2s[15]	&134M &- &- &- &- &333.7 &425.3 \\
  Zhang et al.[13] &21.4M &181.8 &277.7 &32.0 &49.8 &467.0 &498.5\\
  MCNN[14] &0.13M &110.2 &173.2 &26.4 &41.3 &377.6 &509.1\\
  FCN[19] &0.32M &126.5 &173.5 &23.8 &33.1 &338.6 &424.5\\
  Switch-CNN[17] &15.1M &90.4 &135.0 &21.6 &33.4 &318.1 &439.2\\
  ResNet-14 &\bf{0.028M} &93.8 &137.1 &\textcolor{red}{{18.1}} &\textcolor{red}{{27.8}} &370.3 &468.1\\
  DR-ResNet &\textcolor{red}{{0.028M}} &\textcolor{red}{{86.3}} &\textcolor{red}{{124.2}} &\bf{14.5} &\bf{21.0} &\textcolor{red}{{307.4}} &\textcolor{red}{{421.6}}\\
  CP-CNN[18] &62.9M &\bf{73.6} &\bf{106.4} &20.1 &30.1 &\bf{295.8}	&\bf{320.9}\\
\hline 
\end{tabular}
\label{tab:PPer}
\end{table*}

\subsubsection{ShanghaiTech dataset}
\label{sssec:subhead}
The ShanghaiTech is a large-scale crowd counting dataset. In order to increase diversity of the training set, we used random cropping and horizontal flipping for data augmentation. The paper [14]  did not provide the perspective maps for ShanghaiTech dataset, which limits the generation of density maps. Therefore, we referenced to [14] and used the geometry-adaptive kernels to generate density maps of Part\_A. For Part\_B, the number of people was relatively sparse and head size changed a little, so we fixed the size of Gaussian kernel to 25 with $\sigma$ of 1.5 for generation of density maps. In Table 2, compared to the other methods, the DR-ResNet was more accurate and robust on Part\_B. On the other hand, the CP-CNN achieved better performance than DR-ResNet on Part\_A. It is important to emphasize that the number of parameters in our model was only 0.028 M, which is more than 500 times smaller than in Switch-CNN and more than 2000 times smaller than in CP-CNN.

\subsubsection{The UCF\_CC\_50 dataset}
\label{sssec:subhead}
The UCF\_CC\_50 dataset consists of 50 gray images from the Internet. Due to limited number of images, size and number of heads in the image changed dramatically, the UCF\_CC\_50 is a very challenging dataset[24]. For fair comparison, we followed the processing standards of other approaches and used a 5-fold cross-validation to validate the performance of DR-ResNet. The Table 2 shows that only CP-CNN has higher performance than our DR-ResNet.

\begin{table}[h]
\caption{Comparison of performances of different methods on Beijing BRT dataset. Bold denotes the best results.} 
\centering 
\begin{tabular}{c c c c} 
\hline 
 Method &PARAMS &MAE &MSE\\
\hline 
  MCNN[14] &0.13M &2.24 &3.35\\
  FCN[19] &0.32M &1.74 &2.43\\
  ResNet-14 &0.028M &1.48 &2.22\\
  DR-ResNet	&\bf{0.028M} &\bf{1.39} &\bf{2.00}\\
\hline 
\end{tabular}
\label{tab:PPer}
\end{table}

\subsubsection{Beijing BRT dataset}
\label{sssec:subhead}
The above-mentioned datasets are not specific for traffic transport domain. On the other hand, the crowd counting is important for intelligent transportation. Therefore, we collected new data and made a new dataset named Beijing BRT where the number of heads varied between 1 and 64. All images were taken at the Bus Rapid Transit (BRT) in Beijing. The size of each image was $640\times360$ pixels. The images contained shadows, glare, and sunshine interference, and time span was from morning till night, therefore this dataset is very similar to the datasets used in practical applications. Accordingly, the Beijing BRT can be considered as valuable and representative dataset. In order to generate a reasonable density map, we provided the perspective maps and combined them. For mentioned dataset, we used 720 images for training and 560 images for testing.

Since most of recent state-of-the-art approaches are based on very large models and do not provide source code, thus, we compared our model only with MCNN and FCN. The detailed comparison results are shown in Table 3 and they indicate that the DR-ResNet outperforms the MCNN and FCN. Consequently, our network can adapt better to the shadow, light and sun interference, and it does not require multi-column structure and can also learn different scales of information. Hence, it is more suitable for real-life applications than MCNN and FCN.

\begin{figure}[htb]
\begin{minipage}[b]{0.28\linewidth}
  \centering
  \centerline{\includegraphics[width=1.2cm]{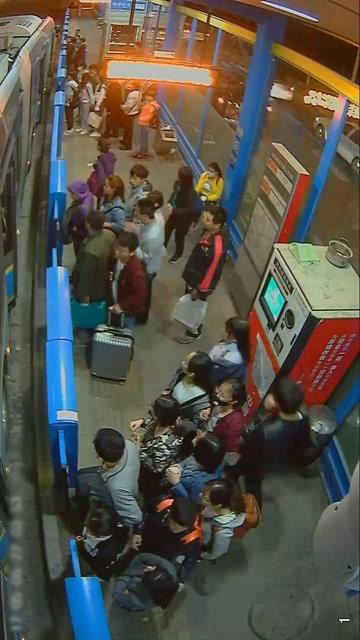}}
  \centerline{(a) Test image}\medskip
\end{minipage}
\hfill
\begin{minipage}[b]{0.28\linewidth}
  \centering
  \centerline{\includegraphics[width=1.2cm]{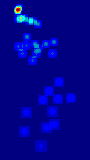}}
  \centerline{(b) Ground-truth:32}\medskip
\end{minipage}
\hfill
\begin{minipage}[b]{0.28\linewidth}
  \centering
  \centerline{\includegraphics[width=1.2cm]{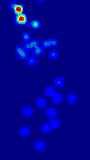}}
  \centerline{(c) Estimation:34}\medskip
\end{minipage}

\begin{minipage}[b]{0.28\linewidth}
  \centering
  \centerline{\includegraphics[width=1.2cm]{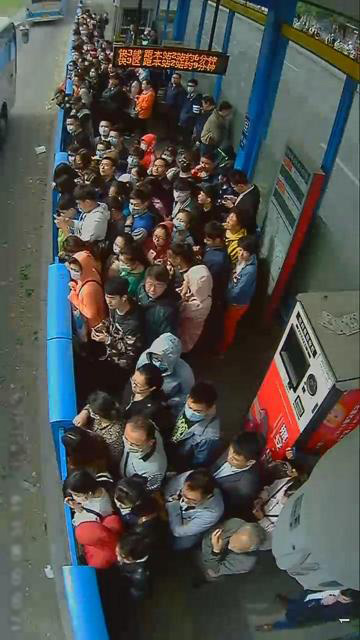}}
  \centerline{(d) Test image}\medskip
\end{minipage}
\hfill
\begin{minipage}[b]{0.28\linewidth}
  \centering
  \centerline{\includegraphics[width=1.2cm]{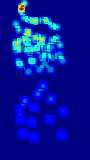}}
  \centerline{(e) Ground-truth:62}\medskip
\end{minipage}
\hfill
\begin{minipage}[b]{0.28\linewidth}
  \centering
  \centerline{\includegraphics[width=1.2cm]{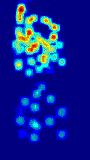}}
  \centerline{(f) Estimation:59}\medskip
\end{minipage}
\caption{The ground truth density maps and estimated density maps of two images of Beijing BRT dataset obtained by DR-ResNet model.}
\label{fig:res}
\end{figure}

\section{CONCLUSIONS}
\label{sec:print}

In this paper, a deeply-recursive convolutional neural network for crowd counting is proposed. Moreover, in order to evaluate the effectiveness of the proposed crowd counting model in the field of intelligent transportation, we introduce a new dataset named Beijing BRT. It is shown that recursive reuse of basic blocks can reduce the number of network parameters while maintaining a good model performance. Experimental results show that with a simple network structure, and small number of parameters and feature layers, the proposed model outperforms most existing models on two key datasets and our Beijing BRT dataset, regardless it has far less number of parameters than other models. Moreover, due to its mentioned characteristics it can be easily applied to embedded devices to quickly calculate the number of people in the image.


\begin{thebibliography}{1}
\bibitem{IEEEhowto:kopka}
Sindagi V A, Patel V M, "A survey of recent advances in CNN-based single image crowd counting and density estimation," \emph{Pattern Recognition Letters}, 2017.
\bibitem{IEEEhowto:kopka}
Dalal, N., Triggs, B., "Histograms of oriented gradients for human detection," \emph{IEEE Conference on Computer Vision and Pattern Recognition}, IEEE. pp. 886-893, 2005.
\bibitem{IEEEhowto:kopka}
Idrees H, Soomro K, Shah M, "Detecting humans in dense crowds using locally-consistent scale prior and global occlusion reasoning," \emph{IEEE transactions on pattern analysis and machine intelligence},  vol. 37, no. 10, pp. 1986-1998, 2015.
\bibitem{IEEEhowto:kopka}
Dollar, P., Wojek, C., Schiele, B., Perona, P., "Pedestrian detection: An evaluation of the state of the art," \emph{IEEE transactions on pattern analysis and machine intelligence}, vol. 34, pp. 743-761, 2012.
\bibitem{IEEEhowto:kopka}
Li, M., Zhang, Z., Huang, K., Tan, T., "Estimating the number of people in crowded scenes by mid based foreground segmentation and head-shoulder detection," \emph{Pattern Recognition (ICPR), 19th International Conference on}, IEEE. pp. 1-4, 2008.
\bibitem{IEEEhowto:kopka}
Chan, A.B., Vasconcelos, N., "Bayesian poisson regression for crowd counting," \emph{IEEE 12th International Conference on Computer Vision}, IEEE. pp. 545-551, 2009.
\bibitem{IEEEhowto:kopka}
Chen K, Loy C C, Gong S, et al, "Feature Mining for Localised Crowd Counting," \emph{British Machine Vision Conference (BMVC)} ,vol. 1, no. 2, p. 3, 2012.
\bibitem{IEEEhowto:kopka}
Ryan, D., Denman, S., Fookes, C., Sridharan, S., "Crowd counting using multiple local features," \emph{Digital Image Computing: Techniques and Applications, DICTA09}, IEEE. pp. 8-88, 2009.
\bibitem{IEEEhowto:kopka}
Lempitsky, V., Zisserman, A., "Learning to count objects in images," \emph{Advances in Neural Information Processing Systems}, pp.1324-1332, 2010.
\bibitem{IEEEhowto:kopka}
Pham, V.Q., Kozakaya, T., Yamaguchi, O., Okada, R., "Count forest: Co-voting uncertain number of targets using random forest for crowd density estimation," \emph{in Proceedings of the IEEE International Conference on Computer Vision}, pp. 3253-3261, 2015.
\bibitem{IEEEhowto:kopka}
Wang, C., Zhang, H., Yang, L., Liu, S., Cao, X., "Deep people counting in extremely dense crowds," \emph{in Proceedings of the 23rd ACM international conference on Multimedia}, ACM. pp. 1299-1302, 2015.
\bibitem{IEEEhowto:kopka}
Fu, M., Xu, P., Li, X., Liu, Q., Ye, M., Zhu, C., "Fast crowd density estimation with convolutional neural networks," \emph{Engineering Applications of Artificial Intelligence}, vol. 43, pp. 81-88, 2015.
\bibitem{IEEEhowto:kopka}
Zhang, C., Li, H., Wang, X., Yang, X., "Cross-scene crowd counting via deep convolutional neural networks," \emph{IEEE Conference on Computer Vision and Pattern Recognition}, pp. 833-841, 2015.
\bibitem{IEEEhowto:kopka}
Zhang, Y., Zhou, D., Chen, S., Gao, S., Ma, Y., "Single image crowd counting via multi-column convolutional neural network," \emph{IEEE Conference on Computer Vision and Pattern Recognition}, pp. 589-597, 2016.
\bibitem{IEEEhowto:kopka}
Onoro-Rubio, D., Lopez-Sastre, R.J., "Towards perspective-free object counting with deep learning," \emph{European Conference on Computer Vision}, Springer. pp. 615-629, 2016.
\bibitem{IEEEhowto:kopka}
Boominathan, L., Kruthiventi, S.S., Babu, R.V., "Crowdnet: A deep convolutional network for dense crowd counting," \emph{Proceedings of the 2016 ACM on Multimedia Conference}, ACM. pp. 640-644, 2016.
\bibitem{IEEEhowto:kopka}
Sam, D.B., Surya, S., Babu, R.V., "Switching convolutional neural network for crowd counting," \emph{IEEE Conference on Computer Vision and Pattern Recognition}, vol. 1, no. 3, p. 6, 2017.
\bibitem{IEEEhowto:kopka}
Sindagi V A, Patel V M. "Generating High-Quality Crowd Density Maps using Contextual Pyramid CNNs," \emph{The IEEE International Conference on Computer Vision (ICCV)}, 2017.
\bibitem{IEEEhowto:kopka}
Marsden, M., McGuiness, K., Little, S., O¡¯Connor, N.E., "Fully convolutional crowd counting on highly congested scenes," arXiv preprint arXiv:1612.00220, 2016.
\bibitem{IEEEhowto:kopka}
Kaiming He, Xiangyu Zhang, Shaoqing Ren, and Jian Sun, "Deep residual learning for image recognition," \emph{IEEE Conference on Computer Vision and Pattern Recognition}, pp. 770-778, 2016.
\bibitem{IEEEhowto:kopka}
Fu, X., Huang, J., Huang, D. Z. Y., Ding, X., Paisley, J, "Removing Rain from Single Images via a Deep Detail Network," \emph{IEEE Conference on Computer Vision and Pattern Recognition}, 2017.
\bibitem{IEEEhowto:kopka}
D. Eigen, J. Rolfe, R. Fergus, and Y. LeCun. "Understanding deep architectures using a recursive convolutional network," \emph{In ICLR Workshop}, 2014.
\bibitem{IEEEhowto:kopka}
Jia, Yangqing, et al, "Caffe: Convolutional architecture for fast feature embedding," \emph{In Proceedings of the 22nd ACM international conference on Multimedia}, pp. 675-678, 2014.
\bibitem{IEEEhowto:kopka}
H. Idrees, I. Saleemi, C. Seibert, and M. Shah, "Multi-source multi-scale counting in extremely dense crowd images," \emph{IEEE Conference on Computer Vision and Pattern Recognition}, pp. 2547-2554, 2013.

\end{thebibliography}

\end{document}